\documentclass{article}
\usepackage{log_2024}						

\usepackage[numbers,compress,sort]{natbib}	

\usepackage[utf8]{inputenc} 
\usepackage[T1]{fontenc}    

\usepackage{url}            
\usepackage{booktabs}       
\usepackage{amsfonts}       
\usepackage{nicefrac}       
\usepackage{microtype}      
\usepackage{xcolor}         

\usepackage{colortbl}
\usepackage{amsmath}
\usepackage{mathtools}
\usepackage{amsthm}
\usepackage{pifont}
\usepackage{relsize}
\usepackage{adjustbox}
\usepackage{wrapfig}
\usepackage{paralist}

\theoremstyle{plain}
\newtheorem{theorem}{Theorem}[section]

\theoremstyle{definition}

\theoremstyle{remark}

\title[CliquePH: Higher-Order Information for Graph Neural Networks]{CliquePH: Higher-Order Information for Graph Neural Networks through Persistent Homology on Clique Graphs}

\author[D. Buffelli et al.]{%
Davide Buffelli\thanks{Equal contribution.}\\
MediaTek Research\\ London, United Kingdom \\
\email{davide.buffelli@mtkresearch.com}\And
Farzin Soleymani\footnotemark[1]\\
Technical University of Munich\\ Munich, Germany\\
\email{fa.soleymani@tum.de}\And
Bastian Rieck\\
University of Fribourg\\ Fribourg, Switzerland\\
\email{bastian.grossenbacher@unifr.ch}
}

\begin{document}

\maketitle

\begin{abstract}
  Graph neural networks have become the default choice by practitioners for graph learning tasks such as graph classification and node classification. Nevertheless, popular graph neural network models still struggle to capture higher-order information, i.e., information that goes \emph{beyond} pairwise interactions. Recent work has shown that persistent homology, a tool from topological data analysis, can enrich graph neural networks with topological information that they otherwise could not capture. Calculating such features is efficient for dimension 0 (connected components) and dimension 1 (cycles). However, when it comes to higher-order structures, it does not scale well, with a complexity of $O(n^d)$, where $n$ is the number of nodes and $d$ is the order of the structures. In this work, we introduce a novel method that extracts information about higher-order structures in the graph while still using the efficient low-dimensional persistent homology algorithm. On standard benchmark datasets, we show that our method can lead to up to $31\%$ improvements in test accuracy.
\end{abstract}

\section{Introduction}
The research area of Graph Neural Networks (GNNs) has received a huge amount of interest in the past few years \cite{waikhom2023survey,wu2020comprehensive}. As a result, GNNs are now the first choice for many graph-related tasks, like graph classification, node classification, and link prediction. Nevertheless, GNNs present some shortcomings when it comes to modeling higher-order structures (i.e., structures in the graph that go beyond pairwise interactions, like \textit{cliques} and \textit{cycles} \cite{diestel2008graph, 10.5555/3134208}). In fact, GNNs follow a message-passing framework \cite{gilmer2017neural} in which each node communicates with its neighbors, but there is no mechanism to go beyond pairwise interactions, which limits the expressiveness of these models \cite{xu2018powerful}.
However, information regarding certain topological structures like cycles and cliques, can be of great importance to graph-related tasks, significantly improving performance \cite{bouritsas2022improving}. 

Topological data analysis (TDA) \cite{wasserman2018topological} is an emerging area of study that uses techniques from topology, and in particular the tool of persistent homology, to study the shape of the data at multiple scales. TDA is particularly well-suited for capturing higher-order information in graphs, as it can provide information related to structures of different orders, including cliques and cycles, which are known to be particularly informative for several graph-related tasks \cite{bouritsas2022improving}. 
Recent works have incorporated topological information related to persistent homology into graph learning tasks with great success \cite{Horn22a}. 
These methods are however limited to persistent homology up to dimension 1, due to the high computational cost that is required for higher dimensions. While this already empowers the model with important structural information that GNNs cannot capture, it is still missing a vast amount of higher-order information.

In this paper we introduce CliquePH, a new topological layer
that can be added to any GNN, and that provides information about persistent homology on higher-order structures. Performing  persistent homology up to dimension  $1$ is extremely efficient and scalable, while higher-dimensional persistent homology is prohibitively expensive \cite{Horn22a}. We then devise a strategy that captures higher-order information while still using the efficient low-dimensional persistent homology. In more detail, we first ``lift'' the graph into multiple \textit{clique graphs} describing the connections of higher-order structures, and then apply persistent homology up to dimension 1 to each ``higher-order'' graph. This strategy allows us to use the very efficient (with complexity linear in the number of nodes) persistent homology of dimension 1, while still extracting information from higher-order structures.

The \textbf{contributions} of this paper can be summarized as follows:
\begin{compactitem}
    \item We introduce a topological layer named CliquePH that can be incorporated into any  GNN and that provides information related to persistent homology on higher-order graph structures.
    \item We highlight theoretical results that provide a strong motivation for the use of higher-order persistent homology in graph learning tasks.
    \item We evaluate our method applied to three popular GNNs on standard graph classification benchmarks, and show performance improvements of up to $31\%$.
\end{compactitem}

\section{Preliminaries}
We denote a graph with $G=(V, E)$, where $V$ is the set of nodes, with $\lvert V \lvert = n$, and $E$ is the set of edges. 
We consider attributed graphs, in which every node has a set of $d$ features. 
The features for each node in a graph are contained in a matrix $\mathbf{X} \in \mathbb{R}^{n \times d}$. We use $\mathcal{N}_v$ to indicate the neighbors of node $v$ (i.e., the nodes connected to $v$ by an edge).

\paragraph{Clique Graphs.} 
A $k$-\textit{vertex clique} in a graph $G=(V, E)$ is a subset of vertices such that every two distinct vertices in the clique have an edge between them, i.e.,  $k = \{v_1, v_2, \dots, v_k\}$, with $v_i \in V$, is a $k$-vertex clique if $(v_i, v_j) \in E \text{ } \forall i \neq j; \text{ } i,j = 1, \dots, k$. 
Let $k_1, k_2, \dots k_z$ be the $r$-vertex cliques in the graph $G$, then the $r$-vertex \textit{clique graph} of $G$ is a graph $K^{(r)}(G) = (V_{K^{(r)}}, E_{K^{(r)}})$, with $V_{K^{(r)}} = \{ k_1, k_2, \dots k_z\}$, and $(k_i, k_j) \in E_{K^{(r)}} \text{ if and only if } i \neq j \text{ and } k_i \cap k_j \neq \emptyset$. In other words, the $r$-vertex clique graph $K^{(r)}(G)$ summarizes the structure of the $r$-vertex cliques in the graph $G$. 

\begin{wraptable}{r}{7cm}
\vspace{-4mm}
    \centering
    \caption{Runtime comparison for one epoch of training of exact higher-order persistent homology up to dimension 2 (denoted with ``FullPH''), 
    and our proposed method CliquePH. Results are for the same architecture and hyperparameters trained for one epoch on MNIST dataset. 
    All results are run on NVIDIA A100.}
    \begin{adjustbox}{max width=\linewidth}
    \centering
    \begin{tabular}{lccc}
        \hline
        \multicolumn{1}{c}{\bf Method} & Time (h:m:s/epoch) & \# Iteration/sec & Speedup\\
        \hline
        FullPH & 9:24:14 &  0.05 it/s & 1 \\
        CliquePH & 00:02:57 &  2.42 it/s & $\sim$191\\
        \hline
    \end{tabular}
    \label{tab:compTime}
    \end{adjustbox}
\end{wraptable}

\paragraph{Persistent Homology.}
\emph{Persistent homology}~\citep{Barannikov94, Edelsbrunner10}, a technique for capturing topological features of data, constitutes the foundation of our proposed method.
In contrast to other graph learning techniques, persistent homology permits us to quantify topological features at \emph{different scales}.
The crucial element enabling multi-scale calculations is a \emph{filtration}, i.e., a consistent ordering of the vertices and edges of a graph, defined using a function $f\colon G \to \mathbb{R}$.  
Filtrations can be either obtained from static descriptor functions that measure certain aspects of a graph, such as its number of connected components or its curvature~\citep{OBray21a}, but recent work also demonstrates that it is possible to \emph{learn} filtrations in an end-to-end fashion~\citep{Hofer20a, Horn22a}.
Regardless of the origin of a filtration, the crucial insight is that, since a graph is a discrete object, its topology can only change at a finite number of \emph{critical thresholds} $t_1, \dotsc, t_k$.
Thus, any function $f\colon G \to \mathbb{R}$ that gives rise to a filtration enables us to turn a graph into a sequence of nested subgraphs $G_1 \subseteq G_2 \subseteq \dotsc \subseteq G_k = G$, where each $G_i$ is defined based on the critical points of~$f$ via $G_i := \{ x \in V \cup E \mid f(x) \leq t_i \}$.
Intuitively, each subgraph consists of all elements of the graph~$G$ whose function value is less than or equal to the critical threshold.
Alongside this sequence of subgraphs, we now calculate the \emph{homology groups}.
These groups capture the topological features---connected components, cycles, voids, and higher-order features---of each subgraph.
Persistent homology represents all topological features arising from the filtration in a set of \emph{persistence diagrams}.
Each persistence diagram consists of a multiset of intervals of the form $(c_i, d_i)$ with $c_i, d_i \in \mathbb{R} \cup \{\infty\}$, with $c_i$ denoting the ``creation time''~(in terms of the critical thresholds ) and 
$d_i$ denoting the  ``destruction time.''
Features with an infinite destruction time are called \emph{essential}; these denote topological features that appear at all scales. 
Prior work in leveraging topological features from graphs ignored clique information due to computational issues: while it is possible to work with higher-order topological features arising from cliques, a naive application of the persistent homology scales with $\mathcal{O}(n^d)$, where $n$ denotes the number of vertices in a graph, and~$d$ denotes the order of cliques\footnote{This complexity is due to the number of possible simplices of dimension $d$.} (we refer to this as \textit{exact} higher order persistent homology).
Our proposed method instead \emph{lifts} cliques into individual skeleton graphs, 
and then performs the efficient persistent homology up to dimension 1. In Table \ref{tab:compTime} we show a time comparison for \textit{one epoch} of training, of our proposed method CliquePH with a lifting of the original graph up to 3-vertex clique graphs, and a modified version making use of exact persistent homology up to dimension 2 (dimension 2 corresponds to triangles, i.e., 3-vertex cliques -- hence the comparison). The exact persistent homology computations take about 190 times more than our method CliquePH\footnote{The output of CliquePH is not the same as performing exact higher-order persistent homology; Table \ref{tab:compTime} is meant to show the impracticality of performing the latter.}. 



\begin{figure}[t]
  \centering
  \includegraphics[width=0.9\linewidth]{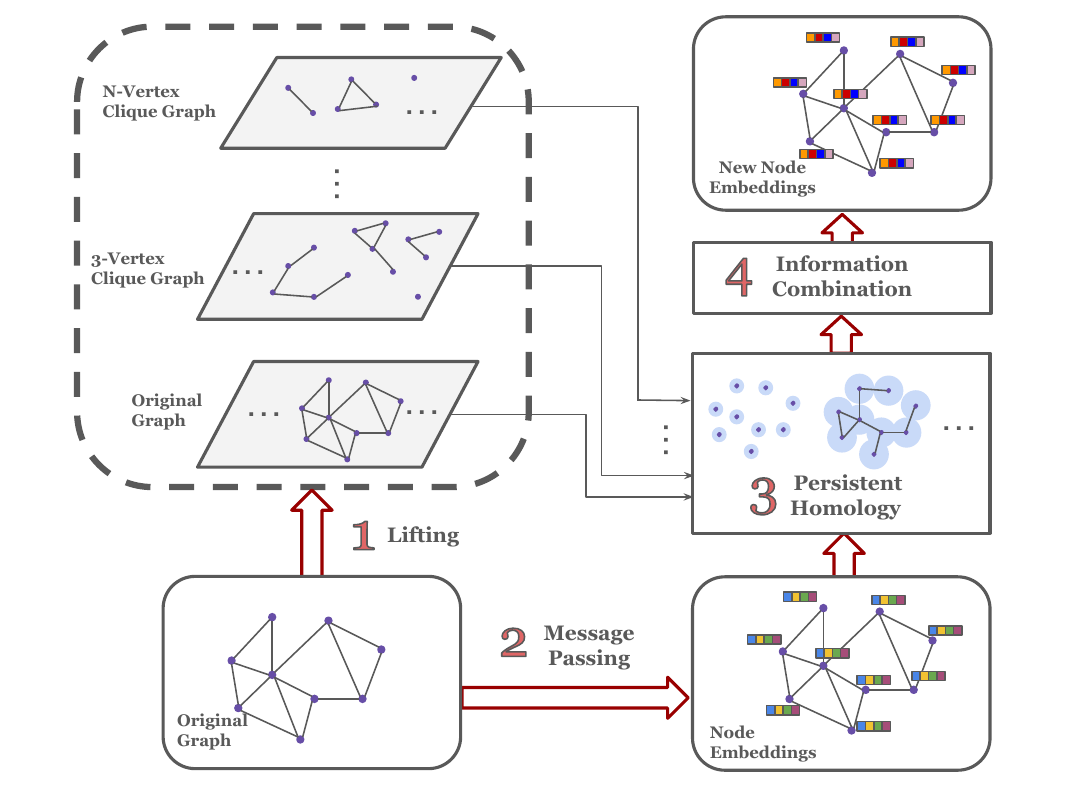}
  \caption{Overview of our method. CliquePH is composed of four stages. (1) First we ``lift'' the original graph by extracting its \textit{clique graphs}. (2) We construct node embeddings using a graph neural network. (3) We use learnable functions to generate filtration values for all nodes and edges in all the lifted graphs. We then perform persistent homology (up to dimension 1) on all lifted graphs. (4) We incorporate the information from persistent homology and message-passing into a single representation. The whole model is trained end-to-end.} \label{fig:cliquePH}
  \vspace{-4mm}
\end{figure}

\paragraph{Graph Neural Networks.} GNNs are deep learning models that operate on graphs. Most popular GNNs adhere to the \textit{message-passing} framework \cite{gilmer2017neural}.
Let $\mathbf{H}^{(\ell)} \in \mathbb{R}^{n \times d^{\prime}}$ be the matrix of node representations at iteration $\ell$ (with $\mathbf{H}^{(0)} = \mathbf{X}$, and $\mathbf{H}^{(i)}_{v}$ used to indicate the representation of node $v$ at iteration $i$). Given a permutation-invariant \textit{aggregation} function $\Phi$, and a learnable \textit{update} function $\Psi_{\theta}$ (usually a neural network), a message passing layer updates  node representations via
\begin{align}
    \mathbf{m}^{(\ell)}_{v} &= \Phi( \{\!\{ \mathbf{H}_{u}^{(\ell)} \text{ } \lvert \text{ } u \in \mathcal{N}_v \}\!\}) \\
    \mathbf{H}_{v}^{(\ell+1)} &= \Psi_{\theta}(\mathbf{H}_{v}^{(\ell)}, \mathbf{m}^{(\ell)}_{v})
\end{align}
where $ \{\!\{ \cdot \}\!\}$ indicates a multiset, and $m_{v}$ is the message received by node $v$ from its neighbours. After $L$ message-passing layers, the node embeddings $\mathbf{H}^{(L)}$ are used to perform a given task (e.g., they are fed to a classifier network), and the whole system is trained end-to-end. For graph-level tasks, it is common to adopt a permutation invariant \textit{readout} function that aggregates all the node representations into a unique graph embedding (e.g., through averaging).

\paragraph{Topological Graph Neural Networks.}
TOGL \cite{Horn22a} is a layer that can be added in-between message passing iterations in GNNs to incorporate global topological information extracted by persistent homology up to dimension 1.
Given the node embedding at layer $\ell$, TOGL uses a learnable filtration function\footnote{TOGL actually uses a set of filtration functions, but we introduce the method with only one for clarity.} $f_{\theta}: \mathbf{H}_{v}^{(\ell)} \rightarrow \mathbb{R}, \forall v \in V$ to assign a filtration value to each node. Filtration values for each edge are then obtained by taking the maximum filtration value between the two nodes connected by the edge. 
Successively, persistent homology up to dimension 1 is computed, returning a persistence diagram for dimension $0$ and one for dimension $1$: $\mathcal{D}^{(0)}, \mathcal{D}^{(1)}$. 
The persistence diagrams are then embedded using DeepSet \cite{zaheer2017deep} networks $S_{\theta}^{(0)}, S_{\theta}^{(1)}$, encoding each diagram into a vector: 
$\mathbf{A}_0 = S_{\theta}^{(0)}(\mathcal{D}^{(0)}), \mathbf{A}_1 = S_{\theta}^{(1)}(\mathcal{D}^{(1)})$. 
The information from the diagrams is then incorporated into node embeddings (for dimension 0), and to the graph embedding (for dimension 1).
All the operations in TOGL are differentiable, and the network is learned end-to-end.

\section{Our Method}
Our proposed method aims at introducing information from persistent homology on higher-order structures into GNNs. 
Our method can be decomposed into four components (as shown in Figure \ref{fig:cliquePH}), which we present below.

\subsection{Graph Lifting}
Starting from the original graph $G$, we perform a ``lifting'' operation to extract its clique graphs:
\[
\text{\textit{lift}}(G) = \{ G, K^{(3)}(G), K^{(4)}(G), \dots,  K^{(r)}(G)\}
\]
the maximum size $r$ for the clique graphs can be defined arbitrarily according to the data at hand (e.g., by taking the value that leads to a clique graph with a number of nodes that is least $5\%$ the number of nodes in the original graph). We start from the $3$-vertex clique graph, as the information about nodes and edges is already present in the original graphs, while the higher dimensions are the ones in which GNNs struggle. This step can be done just once, as pre-processing on all graphs in the dataset.

The motivation behind the use of clique graphs is threefold: (1) several works have proposed methods for efficiently enumerating all cliques in large graphs~\citep{CAZALS2008564,10.1145/3409964.3461800,VASSILEVSKA2009254,10.1145/362342.362367}, (2) cliques are heavily studied in graph theory and have a vast amount of applications (e.g., see \citep{evans2010clique}), and (3) cliques have already been used successfully to improve performance on GNNs \citep{bouritsas2022improving}.

In practice, we use $r=5$ in our experiments as available datasets tend to have a very small number of cliques larger than that. In all considered datasets, we have been able to perform this pre-processing step  up to $r=5$, within 1 hour. For instance, complete preprocessing time on MNIST dataset for up to 3-cliques, up to 4-cliques, and up to 5-cliques takes 36.4 minutes, 58.1 minutes, and 60.7 minutes.

\subsection{Message-Passing}
In this step, we obtain node embeddings for the original graph $G$.
In more detail, we perform $L_{m}$ steps of message passing on $G$. Any GNN could be used for this step, and the output is a matrix of node embeddings $\mathbf{H}^{(L_{m})}$.

\subsection{Learnable Persistent Homology}
Once we have obtained embeddings for the nodes in $G$, we incorporate information from persistent homology for all graphs in $\text{\textit{lift}}(G)$. This step is composed of three sub-components presented below.

\paragraph{1 - Filtration values for nodes and edges.} To perform persistent homology, we first need to compute filtration values for dimension $0$ (nodes), which we indicate with $\mathbf{F}^{(\cdot)}_0$, and dimension $1$ (edges), which we indicate with $\mathbf{F}^{(\cdot)}_1$, in all graphs in the lifted set. We first describe how we obtain filtration values for the original graph $G$, and then for the clique graphs $K^{(\cdot)}(G)$.

\textbf{Original graph $G$.} For dimension $0$, we use a two-layer MLP $f_0$ to map the embedding of each node into  $d_f$ filtration values: $\mathbf{F}^{(G)}_0(v) = f_0(\mathbf{H}_{v}^{(L_{m})})$. 
For dimension $1$, we first obtain a representation $\mathbf{r}_{e_i}$ for each edge $e_i=(u,v)$ by concatenating the representations of the nodes it connects: $\mathbf{r}_{e_i} = \text{cat}(\mathbf{H}_{v}^{(L_{m})}(u), \mathbf{H}_{v}^{(L_{m})}(v))$. 
We then obtain the filtration value for the edge using a 2-layer MLP function $f_1$ as follows: 
\begin{equation}
\label{eq:edge_filtration}
\mathbf{F}^{(G)}_1(e_i) = max(\{\mathbf{F}^{(G)}_0(u), \mathbf{F}^{(G)}_0(v)\}) + f_1(\mathbf{r}_{e_i})
\end{equation} 

\textbf{Clique graphs $K^{(\cdot)}(G)$.} For the lifted graphs, we obtain filtration values in a similar way. 
Let $k_i$ be a node in a clique graph, related to a clique between the nodes $\{u_1, u_2, \dots, u_j\}$, we first obtain the embedding for $k_i$ as $\mathbf{e}_{k_i} = \text{cat}(\mathbf{H}_{v}^{(L_{m})}(u_1), \dots, \mathbf{H}_{v}^{(L_{m})}(u_j))$. We then compute the filtration value as 
\begin{equation}
\mathbf{F}^{K^{(\cdot)}}_0(k_i) = \max(\{\mathbf{F}^{(G)}_0(u_1), \dots, \mathbf{F}^{(G)}_0(u_j)\}) + f_{k^{(\cdot)}_0} (\mathbf{e}_{k_i})
\end{equation}
where $f_{k^{(\cdot)}_0}$ is a two-layer MLP (a separate one for each clique graph).
We then obtain the filtration values for dimension 1 following an analogous procedure to the one used for dimension $1$ on the original graph $G$, using a separate learnable function for each clique graph.

\paragraph{2 - Persistent homology.}
We now have filtration values for all the nodes and edges in all the graphs in the lifted set. We can then perform the efficient persistent homology up to dimension 1 to all the graphs, obtaining two persistence diagrams for each graph in the lifted set: 
$\{ (D_{G}^{(0)}, D_{G}^{(1)}), (D_{K^{(3)}}^{(0)}, D_{K^{(3)}}^{(1)}), \dots, (D_{K^{(r)}}^{(0)}, D_{K^{(r)}}^{(1)}) \}$. The diagrams for dimension $0$ have a number of entries equal to the number of nodes in the respective graph, while the diagrams for dimension $1$ have a number of entries equal to the number of independent cycles in the respective graph. In other words, these diagrams are summarizing the evolution (according to the learned filtration values) of the connected components and cycles in each graph. 

\paragraph{3 - Embed persistence diagrams.}
The diagrams for dimension $0$ of the original graph $G$ are passed to a permutation equivariant (set-to-set) DeepSet network (a separate network for each graph in the lifted set is used), which returns a vector for each node. These vectors are stored in a matrix $\mathbf{E}^{(0)}_{G}$.
The diagrams for dimension $0$ for all clique graphs are passed to a permutation invariant DeepSet network (a separate network for each clique graph is used), which returns a single vector $\mathbf{E}^{(0)}_{K^{(\cdot)}}$ for each clique graph $K^{(\cdot)}(G)$.
The diagrams of dimension $1$ for all graphs are passed through a permutation invariant (set-to-vector) DeepSet network (a separate network for each graph in the lifted set is used) to obtain a unique embedding, which returns a single vector for each graph: $\mathbf{E}^{(1)}_{G}, \mathbf{E}^{(1)}_{K^{(3)}}, \dots, \mathbf{E}^{(1)}_{K^{(r)}}$.

\subsection{Information Combination}\label{sec:inf_combination}
We summarize the information from the persistent homology of each clique graph into a unique vector for each dimension. In more detail, we concatenate the vectors $\mathbf{E}^{(0)}_{K^{(\cdot)}}$ and $\mathbf{E}^{(1)}_{K^{(\cdot)}}$, and pass them through a two-layer MLP to obtain a vector $\mathbf{E}_{K^{(j)}}$ for each dimension $j \ge 3$.
Finally, we combine all the information from the persistent homology computations with the embeddings obtained in the message passing step. This is done by adding the embeddings together:
\begin{align}
    \mathbf{H}^{(L_{m})} &+ \mathbf{E}^{(0)}_{G} + \text{\textsc{scatter}}(\mathbf{E}^{(1)}_{G}) + 
    \text{\textsc{scatter}}(\mathbf{E}_{K^{(3)}}) + \dots + \text{\textsc{scatter}}(\mathbf{E}_{K^{(r)}})
\end{align}
where the function \textsc{scatter} indicates that the embeddings of each clique are added to all the node embeddings of the nodes that form that clique.
More formally, given a $3$-vertex clique $k=(u, v, z)$ with embedding $\mathbf{e}_k$, the operation $\mathbf{H} + \text{\textsc{scatter}}(\mathbf{e}_k)$ is doing the following
\begin{equation}
    \mathbf{H}_v = \mathbf{H}_v + \mathbf{e}_k; \text{  } \mathbf{H}_u = \mathbf{H}_u + \mathbf{e}_k; \text{  } \mathbf{H}_z = \mathbf{H}_z + \mathbf{e}_k
\end{equation}
and leaving the rows of $\mathbf{H}$ related to other nodes unaltered.
Additional round(s) of message passing on $G$ can then be performed before passing the embeddings to the final classifier. In the case of a graph-level task, a standard graph pooling method (e.g., averaging) is used.

Finally, we mention that CliquePH does not affect the permutation equivariance of GNNs, and is fully differentiable (hence allowing end-to-end training). More details can be found in Appendix \ref{app:permutation_equiv_diff}. We futher provide a comparison between CliquePH and TOGL in Appendix \ref{app:diff_togl}.

\subsection{Limitations}
There are three main limitations to our method. Firstly, for very large graphs, it may become difficult computationally to use values of $r$ larger than $3$. While our experiments show that $r=3$ is already enough to provide benefits, this can prevent from exploiting the full potential of CliquePH. Secondly, our method avoids the computational complexity of the exact higher order persistent homology procedure, but it is not equivalent to it, and hence does not have the same theoretical guarantees in terms of expressiveness. Finally, in tasks in which higher-order structures are not informative, CliquePH may not be useful, or may even promote overfitting.

\section{Theoretical Motivation}
The expressivity of GNNs is measured by considering their ability of distinguishing non-isomorphic graphs. This is done by relating GNNs to the Weisfeiler-Lehman algorithm (or WL algorithm) \cite{leman1968reduction}.
We refer to \citet{JMLR:v24:22-0240} for a complete introduction to WL, its variants, and its use in machine learning.
The WL algorithm cannot completely solve the graph isomorphism problem (in fact, there is no known polynomial time solution), but it can still distinguish a large number of graphs \cite{babai1979canonical}. 
Modifications to the WL algorithm have been made to increase its distinguishing power. In particular, some ``higher-order'' variants have been proposed. These variants form a hierarchy of algorithms: 1-WL, 2-WL, ..., $k$-WL, each more expressive than the previous, i.e., $i$-WL can distinguish more graphs than $(i-1)$-WL.
It has been shown that standard message-passing GNNs are \emph{strictly} less powerful than the 1-WL algorithm \cite{xu2018powerful}.
Two recent results, which we provide below, present a strong motivation for using persistent homology to enhance the expressivity of GNNs.
\begin{theorem}[Persistent Homology is at least as expressive as the 1-WL 
- Thm. 2 in \citet{Horn22a}]
\label{thm_1wl}
Persistent homology is at least as expressive as the 1-WL algorithm, i.e. if the 1-WL label sequences for two graphs $G$ and $G^\prime$ diverge, there exists an injective filtration $f$ such that the corresponding
0-dimensional persistence diagrams $D_{G}^{(0)}$ and $D_{G^\prime}^{(0)}$ are not equal.
\end{theorem}
\begin{theorem}[$k$-dimensional Persistent Homology is as expressive as  $k$-WL - Thm. 3 in \citet{rieck2023expressivity}]
\label{thm_kwl}
Given $k$-WL colorings of two graphs $G$ and $G^\prime$ that are different, there exists a filtration of $G$ and $G^\prime$ such that the corresponding persistence diagrams in dimension $k - 1$ or dimension $k$ are different.
\end{theorem}
Theorem \ref{thm_1wl} entails that first-order persistent homology information can already make any GNN \emph{strictly} more expressive than the $1$-WL.
%
Theorem \ref{thm_kwl} is a weaker version of this result; it shows that higher-order persistent homology can in fact match the expressiveness of higher-order variants of the WL algorithm---without requiring an enumeration of all $k$-tuples.

As CliquePH makes use of first-order persistent homology information, it is straightforward to see from Theorem \ref{thm_1wl} that a GNN empowered with CliquePH is strictly more powerful than the 1-WL algorithm. 
While we cannot use Theorem \ref{thm_kwl} to theoretically prove that a version of CliquePH with cliques up to order $k$ leads to the expressiveness of $k$-WL (as CliquePH does not perform exact higher-order persistent homology), we provide below some empirical evidence.

\paragraph{Empirical Validation of Expressiveness.}
We consider 6 datasets of strongly regular graphs~\citep{rieck2023expressivity} (i.e., graphs in which all nodes have the same degree), which are known to be hard to distinguish with the 1-WL test. We then randomly initialize a GCN \cite{kipf2016semi}, a GCN with TOGL \cite{Horn22a}, and a GCN with CliquePH (with a lifting up to 3-vertex clique graphs), and use them to produce graph embeddings. All models are set to have the same number of layers (set to 4), and the internal same dimensionalities (set to 128). We then count the number of \textit{unique} representations (a network that can distinguish all graphs will provide a different representation for each graph). We show results averaged of 10 random seeds in Table \ref{tab:untrained}. CliquePH increases the capability of distinguishing regular graphs, improving also over TOGL, which has an expressivity strictly higher than the 1-WL, empirically showing that persistent homology information from clique graphs provides important performance benefits.

\begin{table}[t]
  \centering
  \caption{%
    We report the test accuracy (following standard practice, we report ROC-AUC on OGBG-molhiv) obtained on standard benchmark datasets when only structural information is considered (i.e., no node attributes are used). We compare standard popular GNNs (\mbox{GCN}, \mbox{GIN}, \mbox{GAT}) alone, and with the addition of TOGL and our proposed method, CliquePH. The highest performing model of each dataset is highlighted with a grey background and the highest performing variant for each architecture in \textbf{bold}. The last column on the right shows the average performance improvement provided by TOGL and CliquePH with respect to the base GNN. 
  }
  \label{tab:no_node_attr}
  \begin{adjustbox}{max width=\linewidth}
  \begin{tabular}{lccccccc||c}
    \hline
    & \multicolumn{7}{c}{Graph classification}\\
    \hline
    \textsc{Method}         
    & \multicolumn{1}{c}{DD}                        
    & \multicolumn{1}{c}{ENZYMES}
    & \multicolumn{1}{c}{MNIST}
    & \multicolumn{1}{c}{PROTEINS} 
    & \multicolumn{1}{c}{IMDB-B}
    & \multicolumn{1}{c}{REDDIT-5K}
    & \multicolumn{1}{c||}{OGBG-molhiv}
    & \multicolumn{1}{c}{Avg. $\uparrow$}
    \\
    \hline
    GCN-4 
    & $\mathlarger{68.0} \pm \mathsmaller{3.6}$ 
    & $\mathlarger{22.0} \pm \mathsmaller{3.3}$ 
    & $\mathlarger{76.2} \pm  \mathsmaller{0.5}$ 
    & $\mathlarger{68.8} \pm \mathsmaller{2.8}$ 
    & $\mathlarger{65.1} \pm \mathsmaller{5.1}$
    & $\mathlarger{54.1} \pm \mathsmaller{0.8}$
    & $\mathlarger{66.4} \pm \mathsmaller{1.8}$
    & $\mathlarger{-}$
    \\
    GCN-3-TOGL-1 
    & $\mathlarger{75.1} \pm \mathsmaller{2.1}$ 
    & $\mathlarger{30.3} \pm \mathsmaller{6.5}$ 
    & \boldmath $\mathlarger{84.8} \pm  \mathsmaller{0.4}$ 
    & \boldmath $\mathlarger{73.8} \pm \mathsmaller{4.3}$ 
    & $\mathlarger{72.5} \pm \mathsmaller{3.0}$
    & $\mathlarger{54.3} \pm \mathsmaller{1.4}$
    & $\mathlarger{69.4} \pm \mathsmaller{1.8}$
    & $\mathlarger{5.7}$
    \\
    GCN-3-CliquePH-1 
    & \boldmath $\mathlarger{76.1} \pm \mathsmaller{1.3}$ 
    & \cellcolor{lightgray} \boldmath $\mathlarger{34.3} \pm \mathsmaller{4.7}$ 
    & $\mathlarger{83.6} \pm  \mathsmaller{0.7}$ 
    & $\mathlarger{72.6} \pm \mathsmaller{1.5}$ 
    & \cellcolor{lightgray} \boldmath $\mathlarger{74.7} \pm \mathsmaller{1.2}$
    & \cellcolor{lightgray} \boldmath $\mathlarger{57.7} \pm \mathsmaller{0.5}$
    & \boldmath $\mathlarger{69.6} \pm \mathsmaller{2.0}$
    & \cellcolor{lightgray} \boldmath $\mathlarger{6.9}$
    
    \\
    \hline
    GIN-4 
    & $\mathlarger{75.6} \pm \mathsmaller{2.8}$ 
    & $\mathlarger{21.3} \pm \mathsmaller{6.5}$ 
    & $\mathlarger{83.4} \pm  \mathsmaller{0.9}$ 
    & \cellcolor{lightgray} \boldmath $\mathlarger{74.6} \pm \mathsmaller{3.1}$ 
    & $\mathlarger{71.1} \pm \mathsmaller{3.7}$
    & $\mathlarger{50.2} \pm \mathsmaller{2.4}$
    & $\mathlarger{68.7} \pm \mathsmaller{0.9}$
    & $\mathlarger{-}$
    \\
    GIN-3-TOGL-1 
    & $\mathlarger{76.2} \pm \mathsmaller{2.4}$ 
    & $\mathlarger{23.7} \pm \mathsmaller{6.9}$ 
    & $\mathlarger{84.4} \pm  \mathsmaller{1.1}$ 
    & $\mathlarger{73.9} \pm \mathsmaller{4.9}$ 
    & \boldmath$\mathlarger{74.4} \pm \mathsmaller{1.6}$
    & $\mathlarger{52.7} \pm \mathsmaller{0.9}$
    & $\mathlarger{65.1} \pm \mathsmaller{6.2}$
    & $\mathlarger{0.8}$
    \\
    GIN-3-CliquePH-1 
    & \cellcolor{lightgray} \boldmath $\mathlarger{77.6} \pm \mathsmaller{1.7}$ 
    & \boldmath $\mathlarger{24.7} \pm \mathsmaller{3.1}$ 
    & \cellcolor{lightgray} \boldmath $\mathlarger{88.9} \pm  \mathsmaller{0.6}$ 
    & $\mathlarger{72.9} \pm \mathsmaller{0.9}$ 
    & $\mathlarger{72.4} \pm \mathsmaller{2.8}$
    & \boldmath $\mathlarger{55.0} \pm \mathsmaller{1.4}$
    & \boldmath $\mathlarger{69.6} \pm \mathsmaller{1.8}$
    & \cellcolor{lightgray} \boldmath $\mathlarger{2.3}$
    \\
    \hline
    GAT-4 
    & $\mathlarger{63.3} \pm \mathsmaller{3.7}$ 
    & $\mathlarger{21.7} \pm \mathsmaller{2.9}$ 
    & $\mathlarger{63.2} \pm \mathsmaller{10.4}$ 
    & $\mathlarger{67.5} \pm \mathsmaller{2.6}$ 
    & $\mathlarger{63.6} \pm \mathsmaller{3.9}$
    & $\mathlarger{48.1} \pm \mathsmaller{1.1}$
    & $\mathlarger{51.8} \pm \mathsmaller{5.6}$
    & $\mathlarger{-}$
    \\
    GAT-3-TOGL-1 
    & \boldmath $\mathlarger{75.7} \pm \mathsmaller{2.1}$ 
    & $\mathlarger{23.5} \pm \mathsmaller{6.1}$ 
    & $\mathlarger{77.2} \pm \mathsmaller{10.5}$ 
    & $\mathlarger{72.4} \pm \mathsmaller{4.6}$ 
    & $\mathlarger{71.2} \pm \mathsmaller{2.4}$
    & $\mathlarger{52.7} \pm \mathsmaller{1.4}$
    & $\mathlarger{68.6} \pm \mathsmaller{1.7}$
    & $\mathlarger{8.8}$
    \\
    GAT-3-CliquePH-1 
    & $\mathlarger{75.2} \pm \mathsmaller{1.9}$ 
    & \boldmath $\mathlarger{27.5} \pm \mathsmaller{4.0}$ 
    & \boldmath $\mathlarger{83.8} \pm  \mathsmaller{0.9}$ 
    & \boldmath $\mathlarger{72.6} \pm \mathsmaller{0.8}$ 
    & \boldmath $\mathlarger{71.5} \pm \mathsmaller{2.6}$
    & \boldmath $\mathlarger{53.7} \pm \mathsmaller{1.0}$
    & \cellcolor{lightgray} \boldmath $\mathlarger{69.9} \pm \mathsmaller{1.7}$
    & \cellcolor{lightgray} \boldmath $\mathlarger{10.7}$
    \\
    \hline
  \end{tabular}
  \end{adjustbox}
\end{table}

\begin{table}[t]
    \caption{Number of strongly regular graphs distinguishable by untrained models. CliquePH can enhance the ability of networks to distinguish ``hard'' non-isomorphic graphs.}
    \label{tab:untrained}
    \begin{center}
    \resizebox{\columnwidth}{!}{%
    \begin{tabular}{lcccccc}
    \hline
    \multicolumn{1}{c}{\bf Dataset }  & \multicolumn{1}{c}{Cubic12} & \multicolumn{1}{c}{Cubic14} & \multicolumn{1}{c}{Cubic16} & \multicolumn{1}{c}{Quartic10} & \multicolumn{1}{c}{Quartic11} & \multicolumn{1}{c}{Quartic12} \\
    \multicolumn{1}{c}{ \# Graphs }  & \multicolumn{1}{c}{85} & \multicolumn{1}{c}{509} & \multicolumn{1}{c}{4060} & \multicolumn{1}{c}{59} & \multicolumn{1}{c}{265} & \multicolumn{1}{c}{1544}
    \\ \hline 
    GCN           & $\mathlarger{78} \pm \mathsmaller{0.0}$ & $\mathlarger{458} \pm \mathsmaller{0.0}$ & $\mathlarger{3604} \pm \mathsmaller{0.0}$ & $\mathlarger{56} \pm \mathsmaller{0.0}$ & $\mathlarger{255} \pm \mathsmaller{0.0}$ & $\mathlarger{1427} \pm \mathsmaller{0.0}$\\
    GCN-TOGL          & $\mathlarger{85} \pm \mathsmaller{0.0}$ & $\mathlarger{507.2} \pm \mathsmaller{1.4}$ & $\mathlarger{4049.4} \pm \mathsmaller{3.7}$ & $\mathlarger{59} \pm \mathsmaller{0.0}$ & $\mathlarger{264} \pm \mathsmaller{0.0}$ & $\mathlarger{1540.1}  \pm \mathsmaller{2.1}$\\
    GCN-CliquePH    & $\mathlarger{85} \pm \mathsmaller{0.0}$ & $\mathlarger{509} \pm \mathsmaller{0.0}$ & $\mathlarger{4057.1} \pm \mathsmaller{0.7}$ & $\mathlarger{59} \pm \mathsmaller{0.0}$ & $\mathlarger{265} \pm \mathsmaller{0.0}$ & $\mathlarger{1542.3} \pm \mathsmaller{2.21}$\\
    \hline
    \end{tabular}
    }
\end{center}
\end{table}

\section{Experiments}

To demonstrate the benefits provided by our method in practical scenarios, we follow prior literature~\cite{Horn22a} and apply our method to three GNNs among the most used by practitioners: GCN \cite{kipf2016semi}, GIN \cite{xu2018powerful}, and GAT \cite{velivckovic2018graph}. We report results for each GNN by itself, for the GNN with the addition of TOGL \cite{Horn22a}, and for the GNN with the addition of our method CliquePH. The comparison with TOGL is particularly useful, as TOGL is a topological method that can be used on top of any GNN (like CliquePH), and it makes use of information coming from persistent homology up to dimension one. This then allows us to analyze the benefit provided by the persistent homology information obtained from the higher-order clique graphs that is present in our method.
Finally, we provide an ablation study to observe the effects of the position of the CliquePH layer in the model architecture.


\paragraph{Experimental Setup and Datasets.}
In order to ensure a fair performance comparison, we take the optimal architectures and hyperparameters for the GNNs and for TOGL from prior work \cite{Horn22a}. 
For CliquePH we use the same architecture for the base GNN of TOGL, and we tune only the learning rate and the maximum dimension of the lifted clique graphs (between 3, 4, and 5) using a random search approach (using values in the interval: $(10^{-6}, 10^{-2})$) by comparing the loss on the validation set. 
Furthermore, the reported results are obtained by averaging over ten independent runs with varying random seeds.
The code to replicate our experiments will be publicly released upon acceptance. We provide statistics about the used models and datasets in Appendix \ref{app:arch_dataset_stats}.
In Tables \ref{tab:no_node_attr} and \ref{tab:benchmarks}, the notation  ``GCN-4'' indicates a GCN architecture with 4 message passing layers, while ``GCN-3-CliquePH-1'' indicates a GCN architecture with 3 message passing layers, and 1 CliquePH layer. We use the analogous notation for the other GNNs, and for networks with TOGL. All experimental results were conducted on a single NVIDIA A100 GPU, and 20 allocated AMD EPYC 7742 64-Core CPUs, with the Adam optimizer, 4 workers, and a batch size of 32 for small datasets and 128 for large datasets.
We use standard graph classification benchmark datasets: DD~\cite{shervashidze2009efficient}, ENZYMES~\cite{schomburg2004brenda}, MNIST~\cite{lecun2010mnist}, PROTEINS~\cite{dobson2003distinguishing}, IMDB-B~\cite{maas2011learning}, Reddit-5k~\cite{yanardag2015deep}, and OGBG-Molhiv~\cite{hu2020ogb}. 
To ensure the same training, evaluation, and data setup, we use the code provided by the authors of TOGL \cite{Horn22a}, and we report results for TOGL and base models from their paper.

\paragraph{Performance on Structure-based Experiments.}
Following prior work \cite{Horn22a}, we perform ``structure-based'' graph classification experiments, in which we assign random node features. This approach removes the information present in the features, and allows us to assess the models' effectiveness when relying only on structural information. Results are shown in Table \ref{tab:no_node_attr}. Firstly, we observe that CliquePH and TOGL significantly improve the performance of standard GNNs. CliquePH in particular leads to the highest average improvement across datasets (right-most column), with an improvement of $6.9\%$ with respect to the base GCN model, $2.3\%$ for the base GIN, and $10.7\%$ for the base GAT. Looking at individual datasets, we then notice that a CliquePH model is the best performing model on 6 out of 7 datasets. Furthermore, it can be seen that in most cases, the results for CliquePH present a lower variance across runs, showing an increased stability with respect to TOGL. 

\paragraph{Performance on Benchmark Datasets.}
We now investigate the effect of CliquePH layer on standard benchmark datasets making use of node attributes. 
We use the same datasets used for the structure-based experiments, except for MNIST, which does not have node attributes. 
Results are shown in Table \ref{tab:benchmarks}. We notice that CliquePH can significantly improve the performance of GNNs even in the presence of rich feature information, while this is not the case for TOGL. In fact, the average performance improvement across datasets for CliquePH is always positive (while for TOGL it is negative for the GIN model) and 
in $90\%$ of the cases, the addition of CliquePH improves the performance of the baseline GNN, with improvements of up to $31\%$. As above, we notice that in the majority of cases, the results for CliquePH have a lower variance than the ones for TOGL, highlighting the increased stability of our method.

\begin{table}[t]
  \centering
  \caption{%
   We report the test accuracy (following standard practice, we report ROC-AUC on OGBG-molhiv) obtained on standard graph classification benchmark datasets (considering node features). We compare standard popular GNNs (\mbox{GCN}, \mbox{GIN}, \mbox{GAT}) alone, and with the addition of TOGL and our proposed method, CliquePH. The highest performing model of each dataset is highlighted with grey background and the highest performing variant for each architecture in \textbf{bold}. The last column on the right shows the average performance improvement provided by TOGL and CliquePH with respect to the base GNN. 
  }
  \label{tab:benchmarks}
  \begin{adjustbox}{max width=\linewidth}
  \begin{tabular}{lccccccc||c}
    \hline
    & \multicolumn{7}{c}{Graph classification}\\
    \hline
    \textsc{Method}         
    & \multicolumn{1}{c}{DD}
    & \multicolumn{1}{c}{ENZYMES}
    & \multicolumn{1}{c}{MNIST}
    & \multicolumn{1}{c}{PROTEINS-FULL}
    & \multicolumn{1}{c}{IMDB-B} 
    & \multicolumn{1}{c}{REDDIT-5K}
    & \multicolumn{1}{c||}{OGBG-molhiv}
    & \multicolumn{1}{c}{Avg. $\uparrow$}
    \\
    \hline
    GCN-4 
    & $\mathlarger{72.8} \pm \mathsmaller{4.1}$
    & \cellcolor{lightgray} \boldmath $\mathlarger{58.3} \pm \mathsmaller{6.1}$
    & $\mathlarger{90.0} \pm \mathsmaller{0.3}$
    & $\mathlarger{76.1} \pm \mathsmaller{2.4}$ 
    & $\mathlarger{68.6} \pm \mathsmaller{4.9}$
    & $\mathlarger{53.7} \pm \mathsmaller{1.7}$
    & $\mathlarger{71.9} \pm \mathsmaller{1.1}$
    & $\mathlarger{-}$
    \\
    GCN-3-TOGL-1 
    & $\mathlarger{73.2} \pm \mathsmaller{4.7}$ 
    & $\mathlarger{53.0} \pm \mathsmaller{9.2}$
    & $\mathlarger{95.5} \pm \mathsmaller{0.2}$
    & $\mathlarger{76.0} \pm \mathsmaller{3.9}$ 
    & \boldmath $\mathlarger{72.8} \pm \mathsmaller{2.3}$ 
    & $\mathlarger{54.5} \pm \mathsmaller{1.2}$ 
    & $\mathlarger{72.6} \pm \mathsmaller{2.0}$
    & $\mathlarger{0.9}$ 
    \\
    GCN-3-CliquePH-1 
    & \boldmath $\mathlarger{75.0} \pm \mathsmaller{1.8}$ 
    & $\mathlarger{55.5} \pm \mathsmaller{4.2}$
    & \boldmath $\mathlarger{95.9} \pm \mathsmaller{0.2}$ 
    & \cellcolor{lightgray} \boldmath $\mathlarger{82.0} \pm \mathsmaller{0.5}$ 
    & $\mathlarger{71.1} \pm \mathsmaller{3.2}$ 
    & \cellcolor{lightgray} \boldmath $\mathlarger{55.6} \pm \mathsmaller{1.2}$ 
    & \boldmath $\mathlarger{75.2} \pm \mathsmaller{2.3}$
    & \cellcolor{lightgray} \boldmath $\mathlarger{2.7}$
    \\
    \hline
    GIN-4 
    & $\mathlarger{70.8} \pm \mathsmaller{3.8}$ 
    & $\mathlarger{50.0} \pm \mathsmaller{12.3}$
    & $\mathlarger{96.1} \pm \mathsmaller{0.3}$
    & $\mathlarger{72.3} \pm \mathsmaller{3.3}$ 
    & $\mathlarger{72.8} \pm \mathsmaller{2.5}$ 
    & $\mathlarger{53.3} \pm  \mathsmaller{1.6}$ 
    & $\mathlarger{69.8} \pm  \mathsmaller{1.1}$ 
    & $\mathlarger{-}$
    \\
    GIN-3-TOGL-1 
    & $\mathlarger{75.2} \pm \mathsmaller{4.2}$ 
    & $\mathlarger{43.8} \pm \mathsmaller{7.9}$
    & $\mathlarger{96.1} \pm \mathsmaller{0.1}$
    & $\mathlarger{73.6} \pm \mathsmaller{4.8}$ 
    & \cellcolor{lightgray} \boldmath $\mathlarger{74.2} \pm \mathsmaller{4.2}$ 
    & $\mathlarger{53.7} \pm  \mathsmaller{1.1}$ 
    & $\mathlarger{67.3} \pm  \mathsmaller{4.0}$ 
    & $\mathlarger{-0.17}$
    \\
    GIN-3-CliquePH-1 
    & \cellcolor{lightgray} \boldmath $\mathlarger{76.1} \pm \mathsmaller{2.6}$ 
    & \boldmath $\mathlarger{54.8} \pm \mathsmaller{8.4}$ 
    & \boldmath $\mathlarger{96.3} \pm \mathsmaller{0.1}$
    & \boldmath $\mathlarger{81.4} \pm \mathsmaller{1.1}$ 
    & $\mathlarger{73.7} \pm \mathsmaller{2.1}$ 
    & \boldmath $\mathlarger{55.1} \pm  \mathsmaller{1.7}$ 
    & \boldmath $\mathlarger{72.9} \pm  \mathsmaller{1.6}$
    & \cellcolor{lightgray} \boldmath $\mathlarger{3.6}$
    \\
    \hline
    GAT-4 
    & $\mathlarger{71.1} \pm \mathsmaller{3.1}$ 
    & $\mathlarger{26.8} \pm \mathsmaller{4.1}$
    & $\mathlarger{94.1} \pm \mathsmaller{0.3}$
    & $\mathlarger{71.3} \pm \mathsmaller{5.4}$ 
    & \boldmath $\mathlarger{73.2} \pm \mathsmaller{4.1}$
    & $\mathlarger{51.4} \pm  \mathsmaller{1.4}$ 
    & $\mathlarger{74.0} \pm  \mathsmaller{2.1}$
    & $\mathlarger{-}$
    \\
    GAT-3-TOGL-1 
    & $\mathlarger{73.7} \pm \mathsmaller{2.9}$ 
    & $\mathlarger{51.5} \pm \mathsmaller{7.3}$
    & $\mathlarger{95.9} \pm \mathsmaller{0.3}$
    & $\mathlarger{75.2} \pm \mathsmaller{3.9}$ 
    & $\mathlarger{70.8} \pm \mathsmaller{8.0}$ 
    & $\mathlarger{52.5} \pm  \mathsmaller{0.9}$ 
    & $\mathlarger{74.7} \pm  \mathsmaller{1.8}$
    & $\mathlarger{4.6}$ 
    \\
    GAT-3-CliquePH-1 
    & \boldmath $\mathlarger{75.4} \pm \mathsmaller{3.1}$ 
    & \boldmath $\mathlarger{57.5} \pm \mathsmaller{9.2}$
    & \cellcolor{lightgray} \boldmath $\mathlarger{96.8} \pm \mathsmaller{0.2}$
    & \boldmath $\mathlarger{80.8} \pm \mathsmaller{1.4}$ 
    & $\mathlarger{70.7} \pm \mathsmaller{3.0}$ 
    & \boldmath $\mathlarger{53.8} \pm  \mathsmaller{1.5}$ 
    & \cellcolor{lightgray} \boldmath $\mathlarger{75.7} \pm  \mathsmaller{1.1}$ 
    & \cellcolor{lightgray} \boldmath $\mathlarger{7.0}$
    \\
    \hline
  \end{tabular}
\end{adjustbox}
\end{table}

\paragraph{Ablation Study.}

We investigate the effects of the position of the CliquePH layer inside a GNN architecture, and the maximum dimension $r$ of the lifted clique graphs. We provide below an overview of the results and we refer to Appendix \ref{app:full_ablation} for the full results. Furthermore, in Appendix \ref{app:deepset} we compare against a version CliquePH with static (non-learnable) embedding functions for the persistence diagrams, instead of (learnable) DeepSet networks, to highlight the benefits provided by the latter. 
We start by considering a 4-layer GNN, with 3 message passing layers and 1 CliquePH layer, and we change the position of the latter. We consider the structure-based scenario to exclude the influence of node features. Results are shown in Figure \ref{fig:ablation_res} (Left) for the DD dataset (other datasets are in the Appendix). We notice that the effects are highly dataset dependent: for DD using CliquePH as the first layer is more effective, while, e.g., for Enzymes the best performance is when the layer is 3rd, similar to PROTEINS.  
We then consider a GNN with a number of layers that vary from 1 to 5 and with CliquePH as the last layer. Results are shown in Figure \ref{fig:ablation_res} (Center). Again we notice that the performance is highly dependent on the dataset, as for DD there is little change in performance for different numbers of layers, while, e.g., for ENZYMES it is important to have 2 to 4 layers prior to CliquePH. On IMDB-B we notice how the addition of CliquePH significantly increases performance even for a 1-layer architecture. Finally, we notice how in \textit{all} cases the addition of CliquePH always leads to performance improvements.
Finally, we apply CliquePH with a maximum dimension of the lifted clique graphs varying from 3 to 5 on the DD dataset. Results are shown in Figure \ref{fig:ablation_res} (Right). We observe that tuning this parameter is important to avoid overfitting. These results suggest that tuning the position of CliquePH, and the value of $r$, is important to obtain the best results, and that CliquePH always leads to performance improvements when no node features are available.


    

\begin{figure}
    \centering
    \minipage{0.30\textwidth}
       \includegraphics[width=1\linewidth]{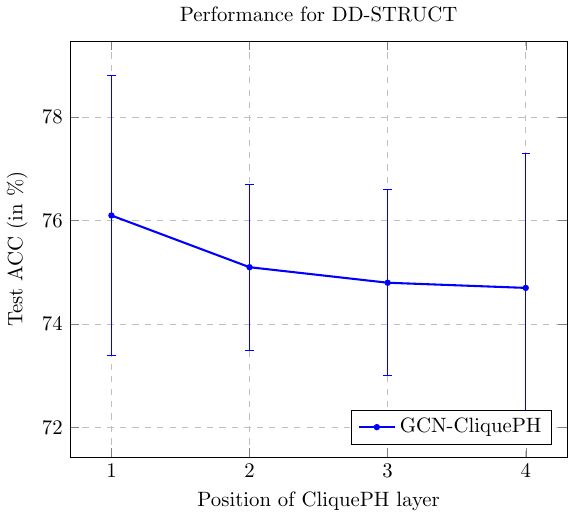}
    \endminipage
    \hfill
    \minipage{0.30\textwidth}
      \includegraphics[width=1\linewidth]{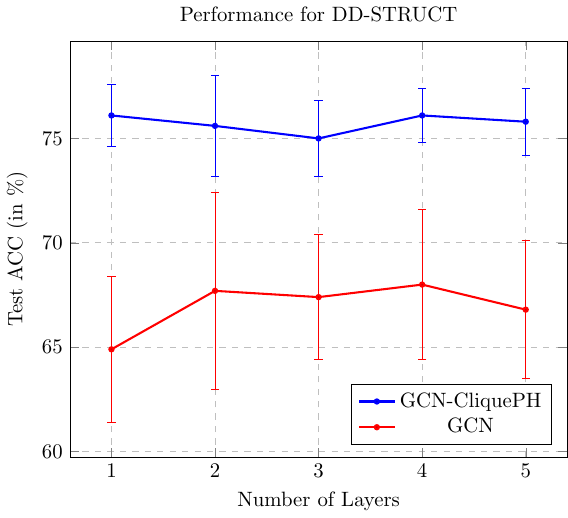}
    \endminipage
    \hfill
    \minipage{0.30\textwidth}
      \includegraphics[width=1\linewidth]{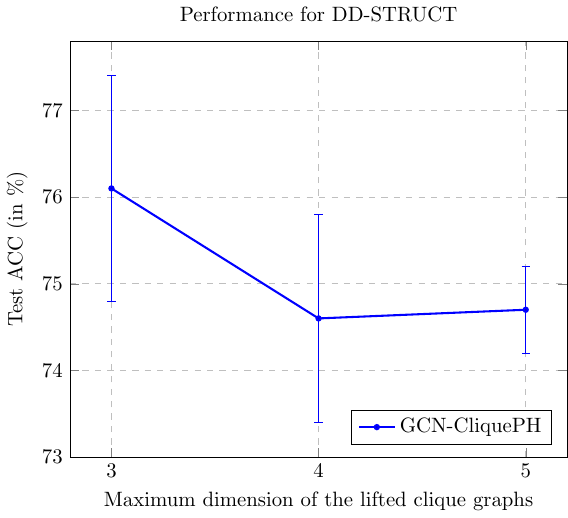}
    \endminipage
    
    \caption{Ablation Results for different architectures. 
    (Left) Test accuracy for the structure-based experiments while changing the position of CliquePH layer. 
    (Center) Test accuracy for the structure-based experiments while increasing the total number of layers in the model. Error bars show standard deviation over 10 runs. 
    (Right) Test accuracy for the structure-based experiments while changing the maximum dimension of the lifted clique graphs of CliquePH.
    }
    \label{fig:ablation_res}
    \vspace{-2mm}
\end{figure}



\section{Related Work}

Graph neural networks and topological data analysis have become very popular and prolific research fields. In this section, we review works at the intersection of graph machine learning and topology, and we refer the interested reader to comprehensive surveys for the latest advancements in graph neural networks \cite{waikhom2023survey,wu2020comprehensive} and topological data analysis \cite{hensel2021survey,chazal2021introduction,leykam2023topological}. 

In the context of \textbf{graph neural networks augmented with topological information}, \citet{bouritsas2022improving} show that adding information about the number of higher-order structures that a node belongs to can improve the theoretical expressivity and practical performance of a model. Similarly, \citet{immonen2023going} use a topological descriptor based on filtrations as additional node features for GNNs. \citet{pmlr-v108-zhao20d} use persistent homology to weight messages in GNNs in the message passing procedure. \citet{he2023sheaf} use sheaf theory to produce node position encodings that improve the performance of graph neural networks on several tasks. \citet{song2023topological} propose a feature augmentation method by obtaining topology embedding of nodes through Node2vec. This introduces topological structure information into an end-to-end model to improve node representation learning.

In addition, several works discuss \textbf{message-passing on topological spaces}. \citet{pmlr-v139-bodnar21a} introduce a model that performs message passing over simplicial complexes. \citet{bodnar2021weisfeiler} successively extended this to cellular complexes. Later work \cite{bodnar2022neural} has also introduced models operating on cellular sheaves. We further mention the work of \citet{hajij2022topological} which provides a unifying framework for deep learning on topological domains.
While these works are related to CliquePH, there are some important differences. Firstly, our method keeps the message passing on the original graph. Secondly, these methods require a specific model, while our method CliquePH can be applied in conjunction with any existing GNN. Thirdly, CliquePH makes use of Persistent Homology, which provides access to information (e.g., the exact number of cliques and cycles in the original graph and in each lifted graph) that is not guaranteed to be accessible to the above methods.

Finally, several works also focus on the explicit connections between \textbf{topological data analysis and graph learning}. \citet{o2021filtration} use graph filtrations to compute efficient graph representations that can outperform those obtained from GNNs. \citet{southern2023curvature} leverage topological tools to robustly evaluate generative graph models. \citet{coupette2023ollivier} introduces the concept of curvature for hypergraphs and show that they can be used to effectively perform clustering. \citet{pmlr-v108-carriere20a} use extended persistent diagrams to produce graph embeddings. \citet{srambical2023filtration} use graph filtrations to classify dynamic graphs. \citet{ye2023treph} and \citet{zhang2022gefl} introduce the use of extended persistent homology, a variant of persistent homology, into graph-related tasks.

\section{Conclusions}
GNNs are powerful models operating on graph structures, but they lack the ability of extracting information about higher-order structures. 
In this paper, we introduce a novel learnable topological layer that provides GNNs with information from persistent homology on higher-order structures present in a graph. Our experimental results show that our method leads to performance improvements of up to $31\%$ on standard benchmarks for the tasks of graph classification.

\section*{Author Contributions}
D.B. and B.R. have come up with the initial idea and formalization. D.B. has coordinated the execution of the project. D.B. has implemented the first version of the code. F.S. has contributed to fixing bugs in the code. F.S. has performed the experiments. F.S. has created the plots and tables for the experimental results. D.B. and B.R. have defined the experimental methodology. D.B. and B.R. have written the paper. F.S. contributed reviewing and editing the paper. B.R. has managed the funding acquisition.

\section*{Acknowledgements}
B.R.\ was partially supported by the Bavarian state government with funds from the \emph{Hightech Agenda Bavaria}. This work has received funding from
the Swiss State Secretariat for Education, Research, and Innovation~(SERI).
F.S acknowledges financial support by The Helmholtz Association of German Research Centres.


\bibliographystyle{unsrtnat}
\bibliography{reference}

\appendix

\section{Permutation Equivariance \& Differentiability} \label{app:permutation_equiv_diff}
\paragraph{Permutation Equivariance.} Most popular GNNs are permutation equivariant (as we do not want the ordering of the nodes to affect the final output), and it is hence important to address whether our proposed method impacts this property. Assuming the use of a permutation equivariant GNN for the message passing operations in CliquPH, and noticing that persistent homology is permutation equivariant by design, we can confirm that CliquePH does indeed preserve this property.

\paragraph{Differentiability.}
The persistent homology computations are differentiable (and hence allow end-to-end training) only with an appropriate choice of function to embed the persistence diagrams. In particular, we refer to the following theorem from \citet[Theorem~1]{Horn22a}.
\begin{theorem}
\label{thm_diff}
Let $f_\theta$ be a vertex filtration function $f_\theta : V \rightarrow R$ with continuous parameters $\theta$, and
let $\Psi$ be a differentiable embedding function (used to embed persistence diagrams) of unspecified dimensions. If the vertex function values
of $f_\theta$ are distinct for a specific set of parameters $\theta^\prime$, i.e. $f_\theta(v) \neq f_\theta(w) \text{ for } v \neq w$, then the map
$\theta \rightarrow \Psi(\text{ph}(G, f_\theta))$ is differentiable at $\theta$.
\end{theorem}
Differentiability thus hinges on unique function values at the vertices of the graph. 
We observe that this condition is always satisfied in practice (as the chances of having all floating point values be exactly the same are infinitesimally small); if need be, it can be enforced by a random perturbation of function values, similar to the well known strategy  of adding random features to GNNs~\citep{ijcai2021p0291,sato2021random}.
Since we use DeepSet (which is a differentiable architecture) as our function to embed the persistence diagrams, we are guaranteed to be respecting Theorem \ref{thm_diff}, thus ensuring the differentiability of our CliquePH method.

\section{Differences Between CliquePH and TOGL} \label{app:diff_togl}
Our method CliquePH is strictly related to TOGL \cite{Horn22a} as both methods provide a ``plug-in'' topological layer for GNNs. Furthermore, both methods rely on a differentiable (and learnable) implementation of persistent homology as the main tool for capturing topological information. 

There are however several differences which we highlight below
\begin{itemize}
    \item CliquePH introduces a lifting operation that allows the model to obtain persistent homology information on the clique graphs representing the  connectivity of higher-order structure in the graph. TOGL on the other hand is strictly limited to information up to dimension 1. This point implies that CliquePH has several functions that are not present in TOGL (e.g., the functions for computing filtration values for the lifted graphs, the functions for embedding higher-order persistence diagrams, the functions for combining information from lifted graphs).
    \item In eq. \ref{eq:edge_filtration}, the  learnable function $f_1$ is not present in the original TOGL layer, and provides CliquePH  with more flexibility. In fact, this provides the model with an additional learnable function that can modify the filtration value of the edge without modifying the values for the nodes connected by that edge.
    \item CliquePH uses an MLP for combining information (section \ref{sec:inf_combination}) from persistence diagrams into node embeddings. TOGL does not use a learnable function for this step.
\end{itemize}

\section{Full Ablation results}\label{app:full_ablation}
\begin{figure}[h]
    \centering
    \minipage{0.31\textwidth}
    \includegraphics[width=1\linewidth]{DD_Clique_Pos.pdf}
    \endminipage
    \minipage{0.31\textwidth}
    \includegraphics[width=1\linewidth]{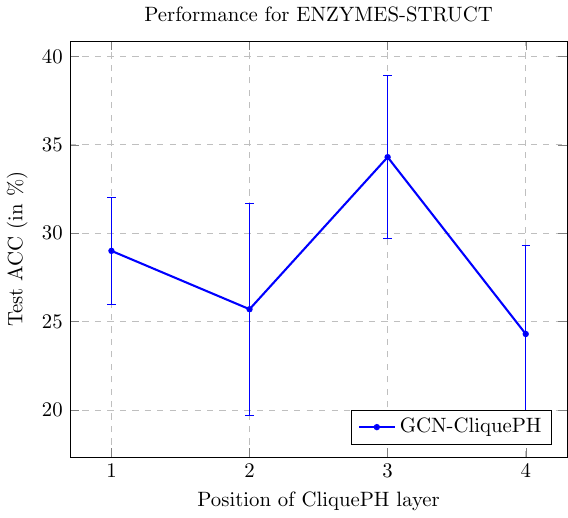}
    \endminipage
    \minipage{0.31\textwidth}
    \includegraphics[width=1\linewidth]{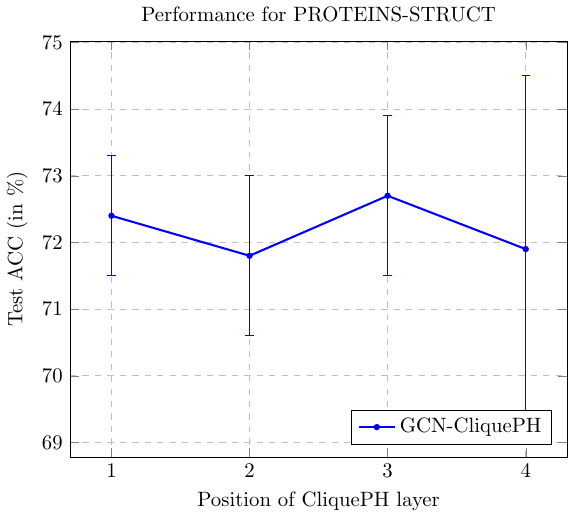}
    \endminipage \\
    \minipage{0.31\textwidth}
    \includegraphics[width=1\linewidth]{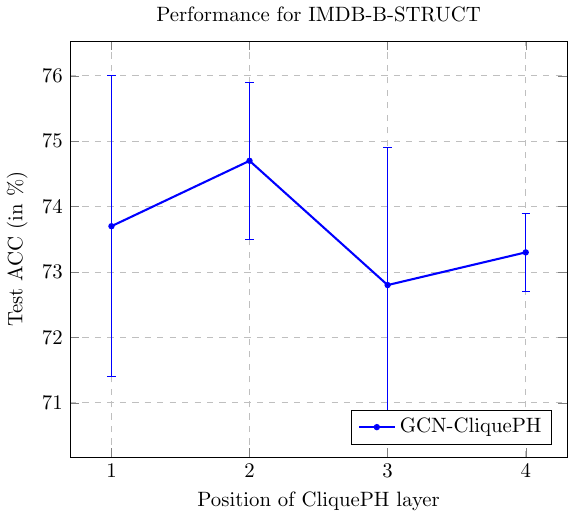}
    \endminipage
    \minipage{0.31\textwidth}
    \includegraphics[width=1\linewidth]{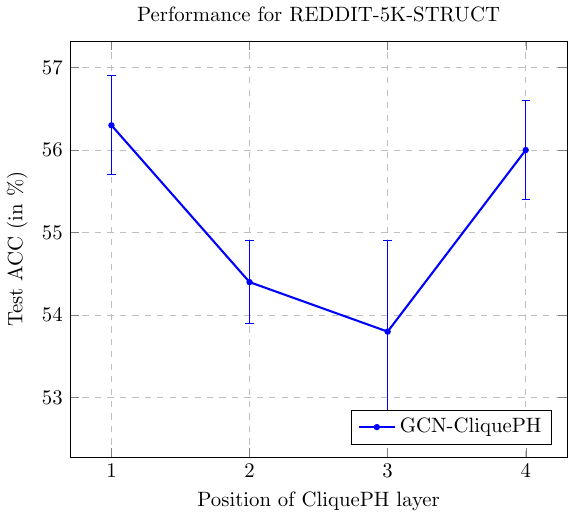}
    \endminipage
    
    \caption{Comparison of test accuracy for the structure-based experiments while changing the position of CliquePH layer in the GCN model architecture. Results averaged over 10 runs. } \label{fig:CliquePos}
\end{figure}

We start by considering a 4-layer GNN, with 3 message passing layers and 1 CliquePH layer, and we change the position of the latter. We consider the structure-based scenario to exclude the influence of node features. Results are shown in Figure \ref{fig:CliquePos} (Center) for the DD, ENZYMES, and PROTEINS datasets. We notice that the effects are highly dataset dependent: for DD using CliquePH as the first layer is more effective, while for Enzymes the best performance is when the layer is  3rd, similar to PROTEINS.  

\begin{figure}[h]
    \centering
    \minipage{0.31\textwidth}
    \includegraphics[width=1\linewidth]{DD_Num_Layers.pdf}
    \endminipage
    \minipage{0.31\textwidth}
    \includegraphics[width=1\linewidth]{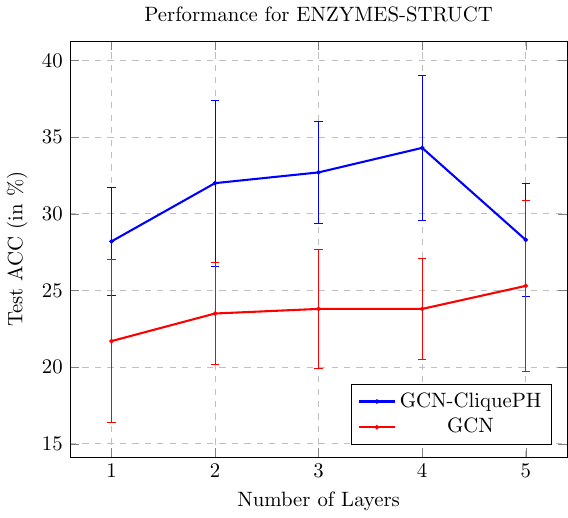}
    \endminipage
    \minipage{0.31\textwidth}
    \includegraphics[width=1\linewidth]{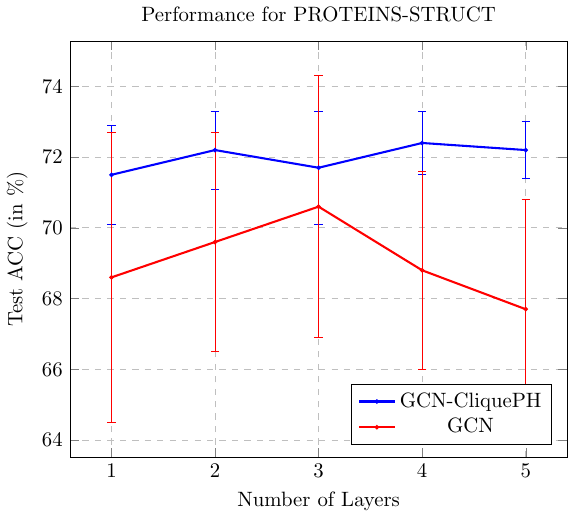}
    \endminipage \\

    \minipage{0.31\textwidth}
    \includegraphics[width=1\linewidth]{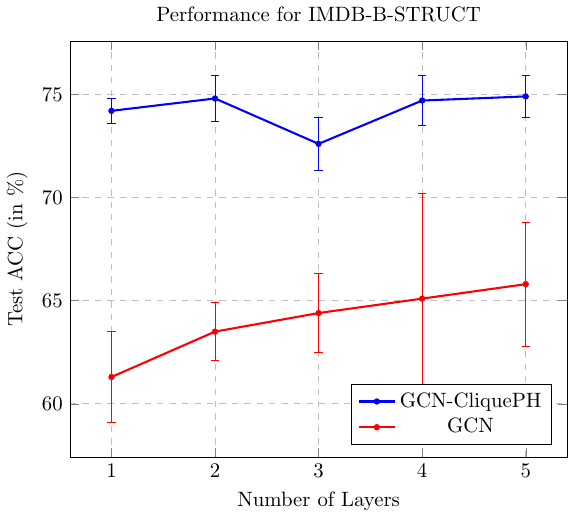}
    \endminipage
    \minipage{0.31\textwidth}
    \includegraphics[width=1\linewidth]{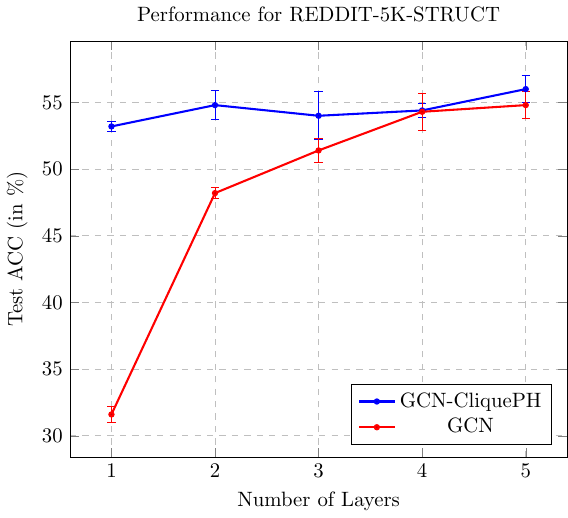}
    \endminipage

    \caption{Comparison of test accuracy for the structure-based experiments while increasing the total number of layers in the model architecture. Error bars show the standard deviation over 10 runs.} \label{fig:NumLayers}
\end{figure}

We then consider a GNN with a number of layers that vary from 1 to 5 and with CliquePH as the last layer. Results are shown in Figure \ref{fig:NumLayers}. Again we notice that the performance is highly dependent on the dataset, as for DD there is little change in performance for different numbers of layers, while for ENZYMES it is important to have 2 to 4 layers prior to CliquePH. On PROTEINS we notice how the addition of CliquePH significantly increases performance even for a 1-layer architecture. Finally, we notice how in \textit{all} cases the addition of CliquePH always improves the final performance.

\begin{figure}[!h]
    \centering
    \minipage{0.31\textwidth}
    \includegraphics[width=1\linewidth]{DD_ablation_num_of_clique_GCN.pdf}
    \endminipage
    \minipage{0.31\textwidth}
    \includegraphics[width=1\linewidth]{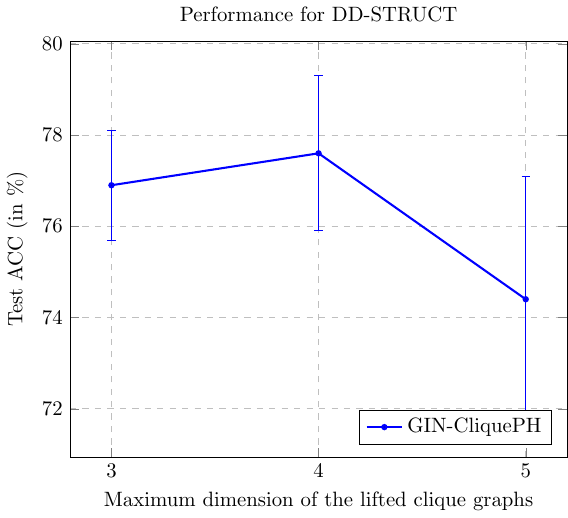}
    \endminipage
    \minipage{0.31\textwidth}
    \includegraphics[width=1\linewidth]{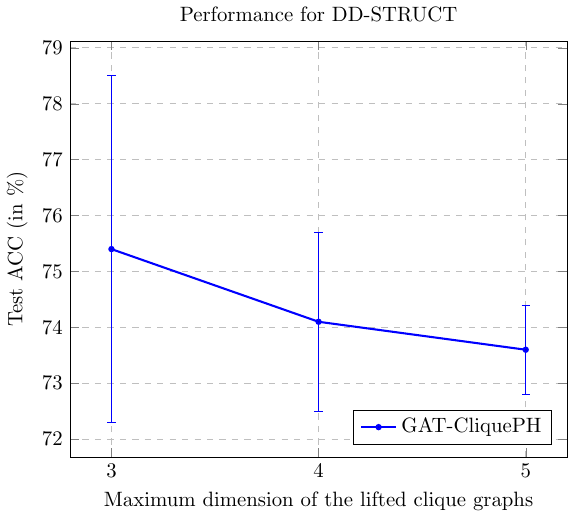}
    \endminipage \\

    \minipage{0.31\textwidth}
    \includegraphics[width=1\linewidth]{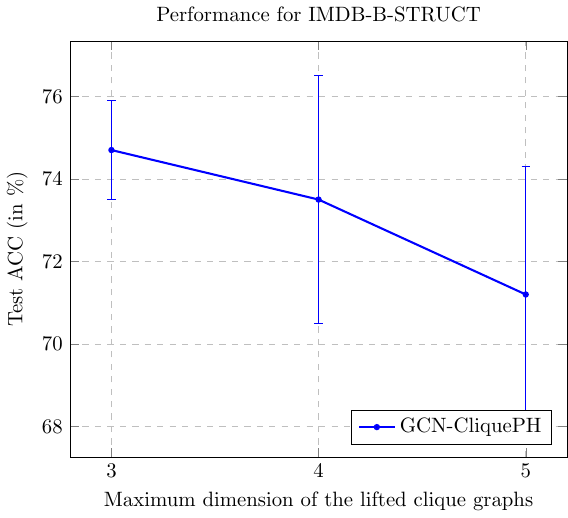}
    \endminipage
    \minipage{0.31\textwidth}
    \includegraphics[width=1\linewidth]{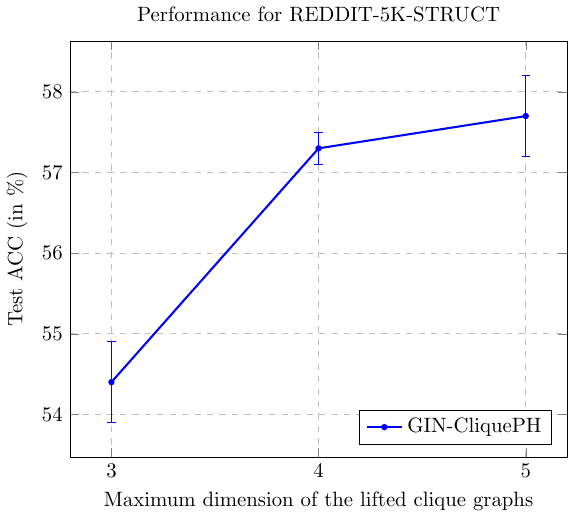}
    \endminipage
    
    \caption{Comparison of test accuracy for the structure-based experiments while changing the maximum dimension of the lifted clique graphs of CliquePH. Results averaged over 10 runs.} \label{fig:maxClique}
\end{figure}

Finally, in Figure \ref{fig:maxClique} we show the performance of a GCN, GIN, and GAT model with different maximum dimension of the lifted graph (varying between 3, 4, and 5) on the DD dataset in a structure-based setting. The results confirm the importance of tuning this parameter to avoid overfitting the data. We also notice that GIN, the theoretically most expressive architecture of the three, achieves the highest performance with a maximum dimension of 4, while GCN and GAT prefer 3.


\section{Learnable vs. Static Embeddings for  Persistence Diagrams} \label{app:deepset}
In this section, we investigate the effect of using a (learnable) DeepSet network instead of static embeddings functions for persistence diagrams. In more details, for the static embeddings we consider the following functions: relational hat \citep{JMLR:v20:18-358}, triangle point transformation, Gaussian point transformation, and line point transformation \citep{pmlr-v108-carriere20a}. 
We report results in Table \ref{tab:deepsets}. We notice that learnable embeddings provide the highest results, and that static embeddings functions can even hurt performance in some datasets.

\begin{table}[t]
  \centering
  \caption{
  We compare two approaches for embedding persistence diagrams: a (learnable) DeepSet model and a method based on static (non-learnable) embeddings. Following standard practice, we report ROC-AUC scores on the OGBG-molhiv dataset using Graph Convolutional Networks (GCN) for evaluation.
  }

  \label{tab:deepsets}
  \begin{adjustbox}{max width=\linewidth}
  \begin{tabular}{lcccccc}
    \hline
    & \multicolumn{6}{c}{Graph classification}\\
    \hline
    \textsc{Method}         
    & \multicolumn{1}{c}{ENZYMES}
    & \multicolumn{1}{c}{PROTEINS-FULL}
    & \multicolumn{1}{c}{IMDB-B}
    & \multicolumn{1}{c}{MNIST}
    & \multicolumn{1}{c}{REDDIT-5K}
    & \multicolumn{1}{c}{OGBG-molhiv}
    \\
    \hline
    GCN-4 
    & $\mathlarger{58.3} \pm \mathsmaller{6.1}$
    & $\mathlarger{76.1} \pm \mathsmaller{2.4}$
    & $\mathlarger{68.6} \pm \mathsmaller{4.9}$
    & $\mathlarger{90.0} \pm \mathsmaller{0.3}$
    & $\mathlarger{53.7} \pm \mathsmaller{1.7}$
    & $\mathlarger{71.9} \pm \mathsmaller{1.1}$
    \\
    GCN-3-CliquePH-1 (Static)
    & $\mathlarger{44.0} \pm \mathsmaller{7.0}$ 
    & $\mathlarger{77.3} \pm \mathsmaller{2.9}$
    & $\mathlarger{70.4} \pm \mathsmaller{4.8}$
    & $\mathlarger{93.8} \pm \mathsmaller{0.5}$
    & $\mathlarger{50.4} \pm \mathsmaller{2.2}$
    & $\mathlarger{74.5} \pm \mathsmaller{1.9}$
    \\
    GCN-3-CliquePH-1 (Ours; DeepSet)
    & $\mathlarger{55.5} \pm \mathsmaller{4.2}$
    & $\mathlarger{82.0} \pm \mathsmaller{0.5}$
    & $\mathlarger{71.1} \pm \mathsmaller{3.2}$
    & $\mathlarger{95.9} \pm \mathsmaller{0.2}$
    & $\mathlarger{55.6} \pm \mathsmaller{1.2}$
    & $\mathlarger{75.2} \pm \mathsmaller{2.3}$
    \\

    \hline
  \end{tabular}
\end{adjustbox}
\end{table}

\section{Node Classification Experiments}
While we believe the benefits of CliquePH are more visible for graph classification tasks (in which higher-order structural information usually plays a much more important role), we performed a small experiment on the Cora-ML \cite{bojchevski2017deep}, Coauthor CS \cite{shchur2018pitfalls}, and Coauthor Physics \cite{shchur2018pitfalls} datasets. We followed the same practices and architectures we used for the graph classification datasets plus an additional 0.2 drop-out on all architectures to prevent overfitting. We adhered to the widely accepted practice of training-validation-test splits of 60\%-20\%-20\%.
We report results in Table \ref{tab:node_class}, and show that CliquePH can provide benefits also for node-level tasks.

\begin{table}[t]
  \centering
  \caption{%
   We report the test accuracy for node classification. We compare standard popular GNNs (\mbox{GCN}, \mbox{GIN}, \mbox{GAT}) alone, and with the addition of TOGL and our proposed method, CliquePH. The highest performing model of each dataset is highlighted with grey background and the highest performing variant for each architecture in \textbf{bold}. 
  }

  \label{tab:node_class}
  \begin{adjustbox}{max width=\linewidth}
  \begin{tabular}{lccc}
    \hline
    \textsc{Method}         
    & \multicolumn{1}{c}{Cora-ML}
    & \multicolumn{1}{c}{Coauthor CS}
    & \multicolumn{1}{c}{Coauthor Physics}
    \\
    \hline
    GCN-4 
    & $\mathlarger{92.4} \pm \mathsmaller{0.8}$
    & $\mathlarger{92.9} \pm \mathsmaller{0.1}$
    & $\mathlarger{96.1} \pm \mathsmaller{0.06}$
    \\
    GCN-3-TOGL-1 
    & $\mathlarger{93.0} \pm \mathsmaller{1.1}$
    & $\mathlarger{93.0} \pm \mathsmaller{0.1}$
    & $\mathlarger{96.4} \pm \mathsmaller{0.06}$
    \\
    GCN-3-CliquePH-1 
    & \boldmath $\mathlarger{93.4} \pm \mathsmaller{0.6}$
    & \boldmath $\mathlarger{93.7} \pm \mathsmaller{0.3}$
    & \cellcolor{lightgray} \boldmath $\mathlarger{96.6} \pm \mathsmaller{0.07}$
    \\
    \hline
    GIN-4 
    & $\mathlarger{92.9} \pm \mathsmaller{1.3}$
    & $\mathlarger{92.6} \pm \mathsmaller{0.2}$
    & $\mathlarger{95.9} \pm \mathsmaller{0.1}$
    \\
    GIN-3-TOGL-1 
    & $\mathlarger{93.2} \pm \mathsmaller{0.6}$
    & $\mathlarger{93.2} \pm \mathsmaller{0.3}$
    & $\mathlarger{96.3} \pm \mathsmaller{0.05}$
    \\
    GIN-3-CliquePH-1 
    & \boldmath $\mathlarger{94.2} \pm \mathsmaller{1.0}$
    & \boldmath $\mathlarger{94.3} \pm \mathsmaller{0.3}$
    & \boldmath $\mathlarger{96.4} \pm \mathsmaller{0.01}$
    \\
    \hline
    GAT-4 
    & $\mathlarger{93.7} \pm \mathsmaller{1.2}$
    & $\mathlarger{93.2} \pm \mathsmaller{0.2}$
    & $\mathlarger{96.2} \pm \mathsmaller{0.2}$
    \\
    GAT-3-TOGL-1 
    & $\mathlarger{93.8} \pm \mathsmaller{0.7}$
    & $\mathlarger{93.3} \pm \mathsmaller{0.4}$
    & $\mathlarger{96.1} \pm \mathsmaller{0.2}$
    \\
    GAT-3-CliquePH-1 
    & \cellcolor{lightgray} \boldmath $\mathlarger{94.4} \pm \mathsmaller{0.8}$
    & \cellcolor{lightgray} \boldmath $\mathlarger{94.3} \pm \mathsmaller{0.3}$
    & \boldmath $\mathlarger{96.4} \pm \mathsmaller{0.07}$
    \\
    \hline
  \end{tabular}
\end{adjustbox}
\end{table}

\section{Architecture \& Datasets Statistics}\label{app:arch_dataset_stats}
We report statistics for the datasets in Table \ref{tab:dataset_stats}. 

In Tables \ref{tab:cliqueph1_gcn3_params_combined}, \ref{tab:togl1_gcn3_params_combined}, \ref{tab:gcn_params}, we report the parameter count for the considered GCN-TOGL, GCN-CliquePH, and base GCN. Notice how the number of parameters remains almost equal between all models. This is due to our choice of replacing a GCN layer with a topological layer (rather than adding a layer on top), and ensures a fairer comparison.

\begin{table}[t]
  \centering
  \caption{Dataset statistics}
  \label{tab:dataset_stats}
  \begin{adjustbox}{max width=\linewidth}
  \begin{tabular}{lcccccc}
    \hline
    & \multicolumn{6}{c}{Graph classification}\\
    \hline
    \textsc{Method}    
    & \multicolumn{1}{c}{\#graphs}
    & \multicolumn{1}{c}{\#nodes}
    & \multicolumn{1}{c}{\#edges}
    & \multicolumn{1}{c}{\#features}
    & \multicolumn{1}{c}{\#classes} 
    & \multicolumn{1}{c}{max clique size}
    \\
    \hline
    DD
    & $\mathlarger{1178}$
    & $\mathlarger{284.32}$
    & $\mathlarger{1431.32}$
    & $\mathlarger{89}$
    & $\mathlarger{2}$
    & $\mathlarger{7}$
    \\
    ENZYMES
    & $\mathlarger{600}$
    & $\mathlarger{32.63}$
    & $\mathlarger{124.27}$
    & $\mathlarger{3}$
    & $\mathlarger{6}$
    & $\mathlarger{5}$
    \\
    MNIST
    & $\mathlarger{55000}$
    & $\mathlarger{70.56}$
    & $\mathlarger{564.50}$
    & $\mathlarger{1}$
    & $\mathlarger{10}$
    & $\mathlarger{8}$
    \\
    PROTEINS
    & $\mathlarger{1113}$
    & $\mathlarger{39.06}$
    & $\mathlarger{145.63}$
    & $\mathlarger{3}$
    & $\mathlarger{2}$
    & $\mathlarger{5}$
    \\
    IMDB-B
    & $\mathlarger{1000}$
    & $\mathlarger{19.77}$
    & $\mathlarger{193.06}$
    & $\mathlarger{0}$
    & $\mathlarger{2}$
    & $\mathlarger{30}$
    \\
    REDDIT-5K
    & $\mathlarger{4999}$
    & $\mathlarger{508.52}$
    & $\mathlarger{1189.75}$
    & $\mathlarger{0}$
    & $\mathlarger{5}$
    & $\mathlarger{6}$
    \\
    OGBG-molhiv
    & $\mathlarger{41127}$
    & $\mathlarger{25.51}$
    & $\mathlarger{54.94}$
    & $\mathlarger{9}$
    & $\mathlarger{2}$
    & $\mathlarger{4}$
    \\
    Coauthor CS
    & $\mathlarger{1}$
    & $\mathlarger{18333.00}$
    & $\mathlarger{163788.00}$
    & $\mathlarger{6805}$
    & $\mathlarger{15}$
    & $\mathlarger{20}$
    \\
    Coauthor Physics
    & $\mathlarger{1}$
    & $\mathlarger{34493.00}$
    & $\mathlarger{495924.00}$
    & $\mathlarger{8415}$
    & $\mathlarger{5}$
    & $\mathlarger{12}$
    \\
    Cora ML
    & $\mathlarger{1}$
    & $\mathlarger{2995.00}$
    & $\mathlarger{16316.00}$
    & $\mathlarger{2879}$
    & $\mathlarger{7}$
    & $\mathlarger{7}$
    \\
    \hline
  \end{tabular}
\end{adjustbox}
\end{table}

\begin{table}[h]
\begin{minipage}{.48\linewidth}
\centering
\caption{GCN-3-CliquePH-1 Model Parameters and Summary}
\label{tab:cliqueph1_gcn3_params_combined}
\resizebox{\linewidth}{!}{%
\begin{tabular}{|l|l|r|}
\hline
\multicolumn{3}{|c|}{\textbf{Model Components}} \\
\hline
\textbf{Name} & \textbf{Type} & \textbf{Params} \\
\hline
Embedding & Linear & 1.2 M \\
Layers & ModuleList & 125 K \\
Classif & Sequential & 13.3 K \\
CliquePH & SimpleSetTopoLayer & 34.5 K \\
\hline
\multicolumn{3}{|c|}{\textbf{Summary}} \\
\hline
\multicolumn{2}{|l|}{Trainable params} & 1.4 M \\
\multicolumn{2}{|l|}{Non-trainable params} & 0 \\
\multicolumn{2}{|l|}{Total params} & 1.4 M \\
\multicolumn{2}{|l|}{Total estimated model params size} & 5.540 MB \\
\hline
\end{tabular}
}
\end{minipage}
\quad
\begin{minipage}{0.48\linewidth}
\centering
\caption{GCN-3-TOGL-1 Model Parameters and Summary}
\label{tab:togl1_gcn3_params_combined}
\resizebox{\linewidth}{!}{%
\begin{tabular}{|l|l|r|}
\hline
\multicolumn{3}{|c|}{\textbf{Model Components}} \\
\hline
\textbf{Name} & \textbf{Type} & \textbf{Params} \\
\hline
Embedding & Linear & 1.2 M \\
Layers & ModuleList & 65.3 K \\
Classif & Sequential & 13.6 K \\
Togl & SimpleSetTopoLayer & 16.1 K \\
\hline
\multicolumn{3}{|c|}{\textbf{Summary}} \\
\hline
\multicolumn{2}{|l|}{Trainable params} & 1.3 M \\
\multicolumn{2}{|l|}{Non-trainable params} & 0 \\
\multicolumn{2}{|l|}{Total params} & 1.3 M \\
\multicolumn{2}{|l|}{Total estimated model params size} & 5.295 MB \\
\hline
\end{tabular}
}
\end{minipage}
\end{table}


\begin{table}[h]
\centering
\caption{GCN Model Parameters and Summary}
\label{tab:gcn_params}
\begin{tabular}{|l|l|r|}
\hline
\multicolumn{3}{|c|}{\textbf{Model Components}} \\
\hline
\textbf{Name} & \textbf{Type} & \textbf{Params} \\
\hline
Embedding & Linear & 1.2 M \\
Layers & ModuleList & 87.0 K \\
Classif & Sequential & 13.6 K \\
\hline
\multicolumn{3}{|c|}{\textbf{Summary}} \\
\hline
\multicolumn{2}{|l|}{Trainable params} & 1.3 M \\
\multicolumn{2}{|l|}{Non-trainable params} & 0 \\
\multicolumn{2}{|l|}{Total params} & 1.3 M \\
\multicolumn{2}{|l|}{Total estimated model params size} & 5.317 MB \\
\hline
\end{tabular}
\end{table}

\end{document}